\documentclass[letterpaper, 10 pt, conference]{ieeeconf}  
\IEEEoverridecommandlockouts
\usepackage{amsmath, amssymb, amsfonts, bm}
\usepackage{graphicx}
\usepackage{cite}
\usepackage{hyperref}
\usepackage{booktabs}
\usepackage{placeins}

\title{Propeller-Assisted Robust 3D Hopping Robot with \\ Hierarchical Force Allocation}
\author{Chuhan Zhang$^{1,2,3}$, Hongbo Zhang$^{1}$, Yanlin Chen$^{1,4}$, Yunxi Tang$^{1,4}$, \\ Yun-Hui Liu$^{1}$, Mingyi Liu$^{2,3}$, and Xiangyu Chu$^{1,4}$
\thanks{This work was supported in part by the Research Grants Council (RGC) of Hong Kong under Grant 14201725, in part by the Multiscale Medical Robotics Centre, AIR@InnoHK, in part by the InnoHK initiative of the Innovation and
Technology Commission of the Hong Kong Special Administrative Region Government via the Hong Kong Centre for Logistics Robotics, and in part by the CUHK T Stone Robotics Institute.}
\thanks{$^{1}$Department of Mechanical and Automation Engineering, The Chinese University of Hong Kong, Hong Kong SAR.
$^{2}$Department of Mechanical Engineering, Guangdong Technion--Israel Institute of Technology, China.
$^{3}$Department of Mechanical Engineering, Technion--Israel Institute of Technology, Haifa, Israel.
$^{4}$Multiscale Medical Robotics Centre, Hong Kong SAR.}%
}

\begin{document}

\maketitle

\begin{abstract}
Monopedal hopping robots are conceptually simple but highly dynamic and inherently unstable. Achieving robust 3D hopping is still difficult because ground reaction forces are available only during the short stance phase, while the robot is underactuated in flight.
A key unresolved issue is how to improve flight-phase control authority. Propeller assistance provides a promising solution, but it requires careful coordination of leg-generated contact forces and propeller thrusts across stance and flight.
This paper presents Pro-OMEGA2, a propeller-assisted 3D monopedal hopping robot with an active 3-RSR parallel leg and a trunk-mounted tri-rotor for auxiliary attitude regulation. To address the force coordination challenge, we propose a Hierarchical Force Allocation (HFA) framework based on a single rigid body (SRB) model. The leg generates the main stance contact wrench, while the tri-rotor provides auxiliary attitude regulation, compensating the residual attitude moment in stance and maintaining attitude during flight.
Real-robot experiments in indoor and outdoor scenarios demonstrate sustained 3D hopping, including terrain transitions and impulsive push recovery, validating robustness under unmodeled contact and external disturbances.
\end{abstract}

\section{INTRODUCTION}

Robust 3D hopping is still a fundamental challenge in legged robotics, particularly for monopedal systems \cite{ref_raibert1984,ref_raibert1986}. Although 3D monopedal hopping has long been demonstrated, reliable hopping in unstructured environments and under strong disturbances is still difficult. Over the past decades, researchers have developed monopedal platforms ranging from series-elastic hoppers for high power density \cite{ref_haldane2016} to untethered 3D monopods for field deployment \cite{ref_leap2017,ref_goat2017} and lightweight 3-RSR (revolute-spherical-revolute) based hoppers \cite{ref_omega,ref_sigma}. The main difficulty is that leg-generated ground reaction forces are available only during the short stance phase, while the robot is underactuated during the flight phase.
\begin{figure}[t]
    \centering
    \includegraphics[width=1.0\columnwidth]{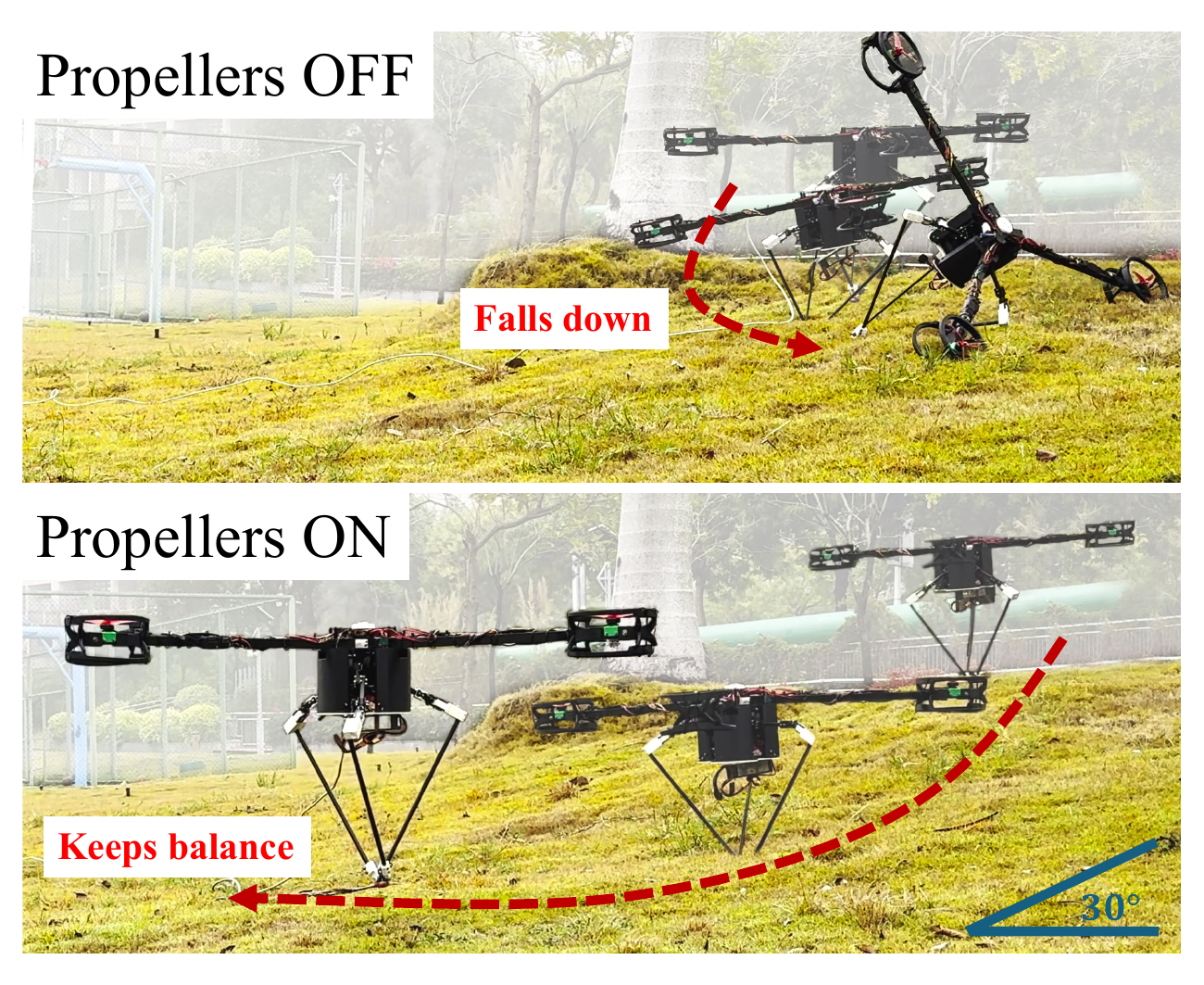}
    \caption{Qualitative comparison on a grassy $30^\circ$ slope. Without propeller assistance the robot becomes unstable; with assistance it remains upright.}
    \label{fig:teaser}
\end{figure}
In conventional monopedal hopping robots, attitude stabilization and disturbance rejection rely primarily on ground reaction forces generated during the stance phase. Beyond monopedal hopping, prior studies in broader legged and underactuated robotic systems have explored inertial appendages to extend attitude control beyond ground contact. Representative examples include tail-assisted stabilization and self-righting \cite{ref_libby2012,ref_changsiu2011,ref_johnson2012}, tail-inspired flight-phase orientation control \cite{ref_chu2019null}, morphable inertial tails for safe landing of falling quadrupeds \cite{ref_tang2023tail}, reaction-wheel-based regulation \cite{ref_machairas2015,ref_lee2023,ref_anzai2021}, and hybrid tail-reaction-wheel designs for spatial reorientation \cite{ref_chu2023tailrw}. These inertial appendages can regulate body attitude during flight or falling phases, but they also have practical drawbacks, including potential collisions between the body and appendages in tail-based designs, added mass and inertia in reaction-wheel systems, and limited direct contribution to stance force generation.

Propeller assistance offers an alternative route for improving monopedal hopping. In monopedal systems, rotor-assisted platforms have demonstrated aerial attitude regulation, perturbation recovery, and terrestrial hopping \cite{ref_pogodrone2022,ref_hopcopter,ref_pogox2024,ref_proomega}. Beyond monopods, the bipedal robot LEONARDO resolves leg-rotor coordination through a joint nonlinear model predictive controller (NMPC) over the full hybrid dynamics \cite{ref_leonardo}. These results suggest that propeller assistance is promising for robust 3D hopping. However, existing propeller-assisted hopping studies do not explicitly resolve how leg-generated contact forces and propeller thrusts should be coordinated across stance and flight to stabilize body attitude without compromising the leg's primary locomotion role. In contrast to a tightly coupled joint optimization such as \cite{ref_leonardo}, we pursue a lightweight hierarchical allocation that keeps the leg primary and uses the tri-rotor only for residual attitude compensation, better suited to the limited onboard computation of a small hopping robot.
\begin{figure*}[t!]
    \centering
    \IfFileExists{figures/design.pdf}{%
        \includegraphics[width=0.95\textwidth]{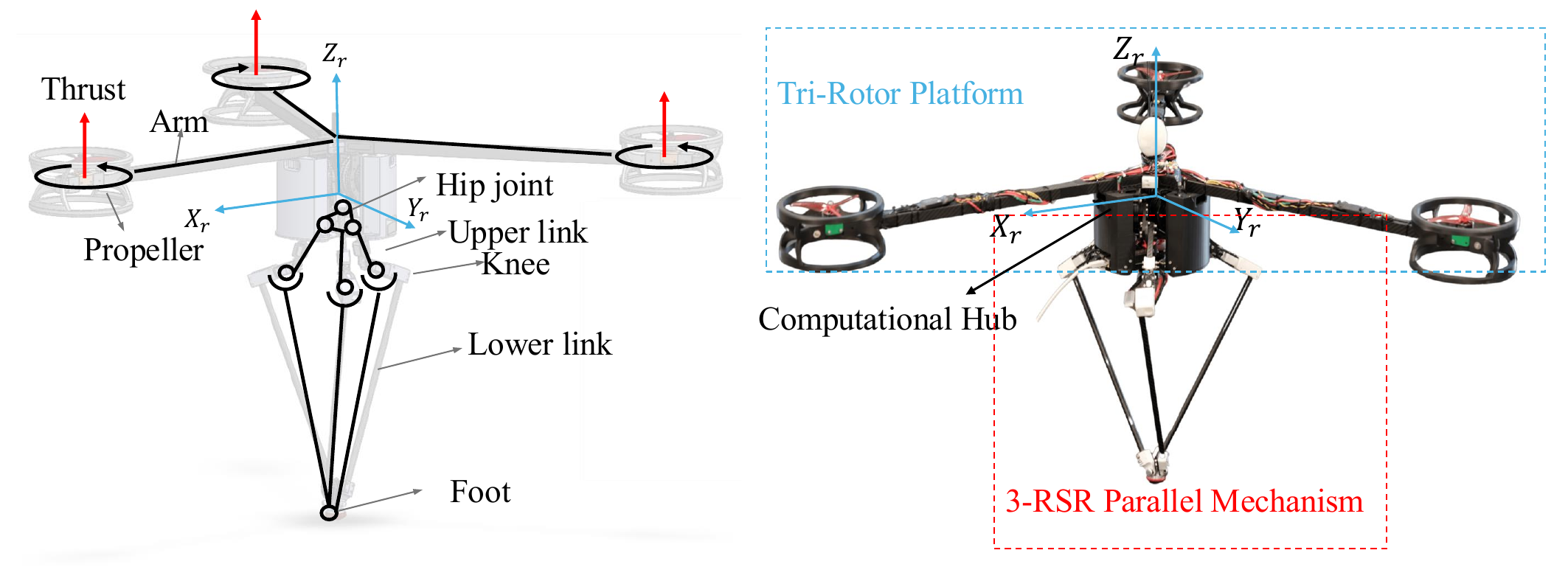}%
    }{%
        \fbox{\parbox{0.95\textwidth}{\centering \small Mechanical Design Placeholder\\(missing: \texttt{figures/design.pdf})}}%
    }
    \caption{Hardware architecture of Pro-OMEGA2. Left: CAD-based decomposition of the tri-rotor, computational hub, and 3-RSR leg. Right: assembled platform with the same modules highlighted.}
    \label{fig:design}
\end{figure*}
To address this issue, we develop Pro-OMEGA2, a 3D hopping robot that combines an active 3-RSR parallel leg with a trunk-mounted tri-rotor unit.
Pro-OMEGA2 extends our group's earlier planar propeller-assisted hopper, Pro-OMEGA \cite{ref_proomega}, to fully spatial 3D operation, where roll, pitch, and horizontal motion are strongly coupled. The leg remains the primary locomotion actuator, while the tri-rotor provides auxiliary attitude regulation and residual attitude compensation.

On this platform, we formulate a Hierarchical Force Allocation (HFA) framework based on a single rigid body (SRB) model that computes a desired centroidal wrench from geometric attitude regulation and an energy-compensated vertical push model, and then allocates it under phase-dependent contact and actuation limits. The leg generates the main stance contact wrench, while the tri-rotor provides auxiliary attitude regulation, compensating the residual attitude moment in stance and maintaining attitude during flight. Extensive real-robot experiments, including outdoor terrain transitions and impulsive push recovery, demonstrate robust 3D hopping under severe disturbances. Fig.~\ref{fig:teaser} qualitatively illustrates the benefit of propeller assistance on a grassy slope.

The primary contributions of this work are the Pro-OMEGA2 hardware platform and the proposed SRB-based HFA framework validated on it.
Specifically, we contribute:

\begin{enumerate}
    \item A hybrid 3D hopping platform, Pro-OMEGA2, which combines an active 3-RSR parallel leg with a trunk-mounted tri-rotor.
    \item An SRB-based HFA controller that allocates feasible stance contact forces and propeller thrusts and uses the tri-rotor as an on-demand residual attitude compensator across contact transitions.
    \item Real-robot validation in indoor and outdoor experiments, including terrain transitions and impulsive push recovery, demonstrating robust 3D hopping under severe perturbations.
\end{enumerate}

\section{SYSTEM AND MODELING}

\subsection{Mechanical Design}

As shown in Fig.~\ref{fig:design}, the Pro-OMEGA2 platform consists of three main structural modules: a 3-RSR parallel leg for ground locomotion, a central computational hub that houses the heavier onboard components, and a tri-rotor platform for aerial attitude control.

The leg module is a 3-RSR spatial parallel mechanism that provides fast omnidirectional leg placement. Specifically, it consists of three identical parallel chains, each comprising an upper link and a lower link. The upper links are attached to the base and are independently driven by three leg motors at the revolute hip joints. Each upper link is connected to its corresponding lower link through a spherical knee joint, and the three lower links connect to the foot link through distal revolute joints that meet at a common point. Driven by the three leg motors, the mechanism enables 3-DoF translational motion of the foot for 3D hopping.

The computational hub is located near the center of the robot and connects the parallel leg to the aerial platform. The batteries and electronic components are packaged inside this hub in a compact layout below the hip level. This arrangement lowers the center of mass and helps reduce rotational inertia.

The tri-rotor platform is mounted above the computational hub in a Y-shaped layout and consists of three propulsion modules. This three-rotor configuration is chosen for its hardware simplicity and low cost, which also helps minimize the upper-body mass. Differential thrust from the three rotors provides roll and pitch control throughout the gait cycle. In the implemented controller, the tri-rotor is used primarily for auxiliary attitude regulation and residual attitude compensation, while the leg remains the main actuator for hopping and horizontal motion. The total thrust-to-weight ratio is kept below unity so that locomotion remains primarily legged rather than aerial.
\begin{figure}[t!]
    \centering
    \IfFileExists{figures/Electronic.pdf}{%
        \includegraphics[width=0.82\columnwidth]{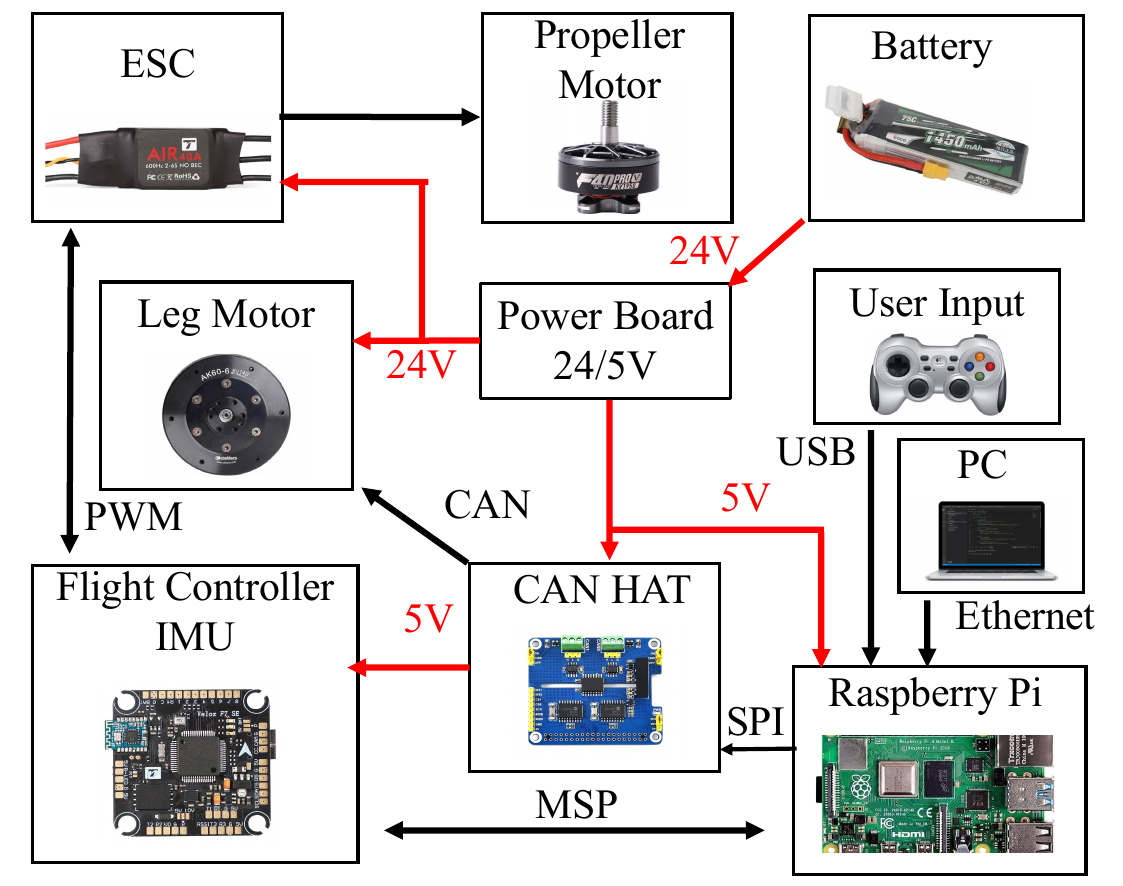}%
    }{%
        \fbox{\parbox{0.9\columnwidth}{\centering \small Electronic system diagram placeholder\\(missing: \texttt{figures/Electronic.pdf})}}%
    }
    \caption{Electronic architecture}
    \label{fig:electronic}
\end{figure}
\subsection{Electronic System Architecture}

As shown in Fig.~\ref{fig:electronic}, the electronic system is organized around a two-layer control architecture. A high-level controller running on an external PC executes the hopping state machine, state estimation, and HFA logic, and communicates with the onboard Raspberry Pi via LCM over Ethernet. The Raspberry Pi implements the hardware interface layer: it sends torque commands to the three leg motors via CAN and exchanges thrust commands and Inertial Measurement Unit (IMU) measurements with a VeloxF7 flight controller via MSP. The flight controller generates PWM signals to the Electronic Speed Controllers (ESCs) that drive the three propeller motors. The measured round-trip latency of the LCM-over-Ethernet link is about $0.2$\,ms on average and below $0.5$\,ms for $99\%$ of cycles, well within the $2$\,ms control period of the $500$\,Hz loop, so it does not noticeably affect the closed loop.

\subsection{SRB Dynamics and Actuation Mapping}

The 3-RSR parallel mechanism exhibits highly non-linear kinematics. However, the leg weighs about $0.35$\,kg, only $9.2\%$ of the $3.75$\,kg total, so its moving mass is small and the dominant dynamics can be reasonably captured by an SRB model \cite{ref_dicarlo2018}. The tri-rotor assembly is rigidly attached to the trunk and moves with it, so the leg is the only mass that moves relative to the body; its small mass fraction keeps the centroidal inertia variation between the extended stance and retracted flight configurations limited. As illustrated in Fig.~\ref{fig:srb}, this abstraction decouples high-level locomotion control from low-level joint non-linearities. We therefore model the body dynamics at the centroidal wrench level and explicitly describe the kinematic/force mappings that connect the SRB wrench to the 3-RSR leg torques and tri-rotor thrusts.

\subsubsection{SRB Wrench Dynamics}
\begin{figure}[t!]
    \centering
    \IfFileExists{figures/srb.pdf}{%
        \includegraphics[width=0.82\columnwidth]{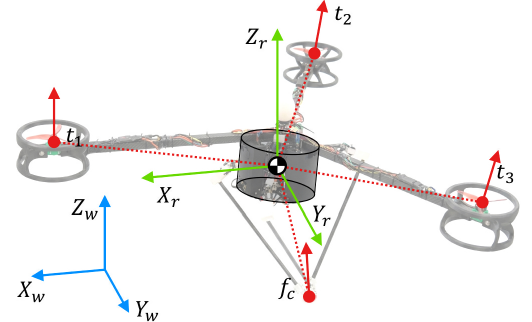}%
    }{%
        \fbox{\parbox{0.9\columnwidth}{\centering \small SRB figure placeholder\\(missing: \texttt{figures/srb.pdf})}}%
    }
    \caption{SRB abstraction with actuation from the stance contact and three propellers.}
    \label{fig:srb}
\end{figure}

To facilitate modeling, we define a body-fixed frame $\{B\}$ at the CoM and a world frame $\{W\}$. Their vertical axes are denoted by $Z_b$ and $Z_w$ in Fig.~\ref{fig:srb}, where $Z_w$ defines the world vertical and gravity acts along $-Z_w$. The rotation matrix $\bm{R}_{WB} \in \mathrm{SO}(3)$ (abbreviated as $\bm{R}$) maps vectors from $\{B\}$ to $\{W\}$. Let $\bm{e}_3 = [0, 0, 1]^\top$ denote the unit $z$-axis of $\{B\}$.
Let $\bm{p}, \bm{v}, \bm{a} \in \mathbb{R}^3$ be the position, velocity, and acceleration of the CoM in the world frame. The net wrench (force $\bm{F}_{net}$ and moment $\bm{\tau}_{net}$) acting on the CoM is generated by the ground reaction force $\bm{f}_c$ and the three propeller thrusts $t_i$:
\begin{equation}
\begin{bmatrix} \bm{F}_{net} \\ \bm{\tau}_{net} \end{bmatrix} = 
\begin{bmatrix} \bm{f}_c \\ \bm{r}_c \times \bm{f}_c \end{bmatrix} + 
\sum_{i=1}^3 \begin{bmatrix} \bm{d}_t \\ \bm{r}_i \times \bm{d}_t \end{bmatrix} t_i,
\label{eq:srb_wrench}
\end{equation}
where $\bm{d}_t = \bm{R}\bm{e}_3$ is the rotor thrust direction expressed in the world frame, and $\bm{r}_c, \bm{r}_i \in \mathbb{R}^3$ are the moment arms from the CoM to the contact point and rotor $i$, respectively, expressed in the world frame.
The system dynamics are then governed by the Newton-Euler equations:
\begin{align}
m\bm{a} &= \bm{F}_{net} + m\bm{g}, \label{eq:srb_trans} \\
\dot{\bm{\omega}} &= \bm{I}^{-1}(\bm{R}^\top\bm{\tau}_{net} - \bm{\omega} \times \bm{I}\bm{\omega}), \label{eq:srb_rot}
\end{align}
where $m$ is the total mass, $\bm{g}$ is the gravity vector, $\bm{\omega}$ is the body angular velocity, and $\bm{I} \in \mathbb{R}^{3 \times 3}$ is the centroidal inertia tensor.

For control synthesis, the objective is to determine a desired wrench $[\bm{F}_{des}^\top, \bm{\tau}_{des}^\top]^\top$ to stabilize the dynamics in Eqs.~\eqref{eq:srb_trans} and \eqref{eq:srb_rot}, and subsequently allocate this wrench to $\bm{f}_c$ and $t_i$ via Eq.~\eqref{eq:srb_wrench} \cite{ref_pucci2017}. The detailed phase-dependent allocation strategy is presented in Section~III.

\subsubsection{Leg Kinematics and Force-to-Torque Mapping}

\begin{figure*}[t]
    \centering
    \includegraphics[width=0.95\textwidth]{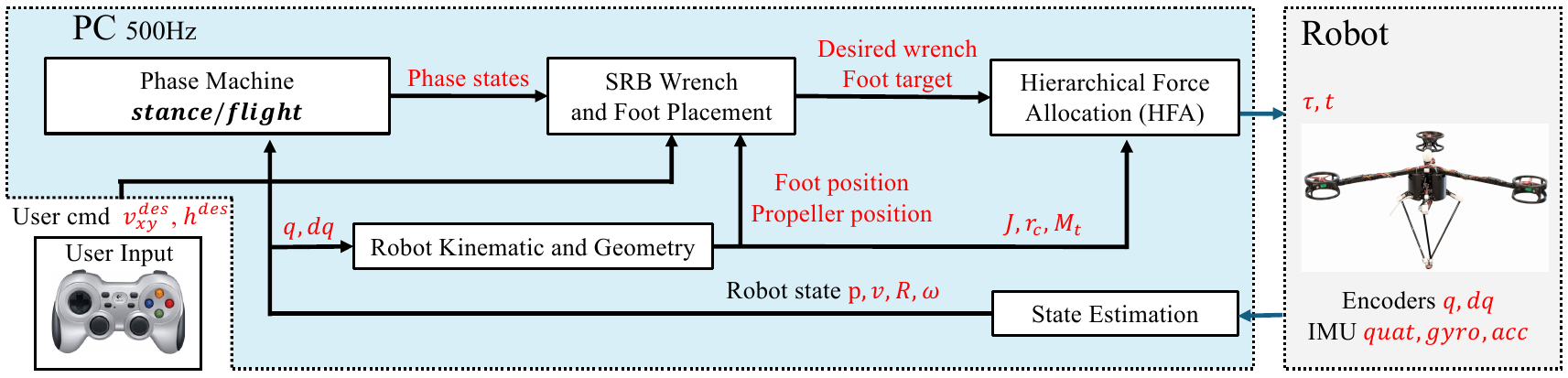}
    \caption{Implemented 500\,Hz PC-side control architecture of the proposed system, from user commands and onboard feedback to leg-torque and tri-rotor-thrust commands.}
    \label{fig:control}
\end{figure*}

The hopping dynamics exhibit two distinct phases. During the flight phase, the 3-RSR leg is unloaded and is used to place the foot for the next touchdown. Denoting the foot position in the body frame by $\bm{p}_{foot}$ and the leg joint coordinates by $\bm{q} = [q_1, q_2, q_3]^\top \in \mathbb{R}^3$, the desired touchdown target is first converted to joint coordinates through the inverse kinematics of the 3-RSR mechanism. Around the resulting configuration, the differential kinematics are described by
\begin{equation}
\dot{\bm{p}}_{foot} = \bm{J}(\bm{q})\dot{\bm{q}},
\end{equation}
where $\bm{J} \in \mathbb{R}^{3 \times 3}$ is the leg Jacobian matrix in the body frame. During the stance phase, the ground reaction force on the body is $\bm{f}_c \in \mathbb{R}^3$ in the world frame. The equal-and-opposite force transmitted through the leg is therefore $\bm{f}_{leg} = -\bm{R}^\top \bm{f}_c$ in the body frame, and the corresponding motor torques are
\begin{equation}
\bm{\tau}_m = \bm{J}^\top \bm{f}_{leg} = -\bm{J}^\top \bm{R}^\top \bm{f}_c,
\label{eq:leg_mapping}
\end{equation}
which follows from the 3-RSR force-transmission structure. The motor torques satisfy $\|\bm{\tau}_m\|_\infty \le \tau_{\mathrm{max}}$, where $\tau_{\mathrm{max}}$ is the per-motor torque limit.
\begin{table}[t!]
\centering
\caption{Physical parameters, thresholds, limits, and high-level gains used in all experiments}
\label{tab:parameters}
\begin{tabular}{lc}
\toprule
Parameter & Value \\
\midrule
\multicolumn{2}{l}{\emph{Physical and phase parameters}} \\
Total Mass $m$ & 3.75 kg \\
Nominal Leg Length $l_0$ & 0.464 m \\
Desired Apex CoM Height $h_{des}$ & 0.664 m \\
Touchdown Threshold $\Delta_{td}$ & 0.020 m \\
\addlinespace[2pt]
\multicolumn{2}{l}{\emph{Actuator and contact limits}} \\
Max Torque $\tau_{\mathrm{max}}$ (per motor) & 20.0 Nm \\
Max Thrust $t_{\mathrm{max}}$ (per rotor) & 10.0 N \\
Max Total Thrust Ratio $T_{\mathrm{max}}/mg$ & 0.82 \\
Baseline Thrust Ratio $\rho$ & 0.02 \\
Friction Coefficient $\mu$ & 0.4 \\
\addlinespace[2pt]
\multicolumn{2}{l}{\emph{Stance-phase gains}} \\
Vertical height gain $k_z$ & 1100 \\
Vertical damping gain $b_z$ & 20 \\
Energy feedback gain $k_E$ & 12 \\
Attitude gains $K_R^{roll,pitch}$ & 100 \\
Attitude damping gains $K_\omega^{roll,pitch}$ & 1 \\
Max roll/pitch moment $\tau_{\mathrm{rp,max}}^{\mathrm{stance}}$ & 30.0 Nm \\
\addlinespace[2pt]
\multicolumn{2}{l}{\emph{Flight-phase gains}} \\
Attitude gains $K_R^{roll,pitch}$ & 60 \\
Attitude damping gains $K_\omega^{roll,pitch}$ & 40 \\
Max roll/pitch moment $\tau_{\mathrm{rp,max}}^{\mathrm{flight}}$ & 25.0 Nm \\
Foot-placement gains $k_v, k_r$ & 0.16, 0.14 \\
\bottomrule
\end{tabular}
\end{table}

The tri-rotor unit provides persistent attitude control throughout both phases. The three propellers are arranged in a Y-shaped layout, with individual thrusts $t_i \in [0, t_{\mathrm{max}}]$ collected as $\bm{t} = [t_1,t_2,t_3]^\top$, where $t_{\mathrm{max}}$ is the per-rotor thrust limit. For robust attitude control, we maintain a nonzero nominal collective thrust $T_{base} = \rho mg$, where $\rho$ denotes the baseline thrust ratio. This keeps the propellers spinning away from the near-zero thrust regime, reducing spin-up delays and helping maintain roll/pitch control through differential thrust. We distribute the bias as $\bm{t} = \tfrac{T_{base}}{3}\mathbf{1}_3 + \delta\bm{t}$, where $\mathbf{1}_3 = [1,1,1]^\top$ and $\delta\bm{t} = [\delta t_1,\delta t_2,\delta t_3]^\top$.

\section{CONTROL SYNTHESIS}

Fig.~\ref{fig:control} summarizes the implemented PC-side control architecture. The control architecture follows a hierarchical structure: an event-driven state machine schedules the hybrid contact phases, a template-level planner generates desired centroidal trajectories, and a two-stage SRB-based allocator computes the commanded contact forces and propeller thrusts.
The main physical parameters, phase thresholds, actuator limits, and high-level controller gains used in all experiments are summarized in Table~\ref{tab:parameters}.
\begin{figure}[t]
    \centering
    \IfFileExists{figures/phase.pdf}{%
        \includegraphics[width=0.82\columnwidth]{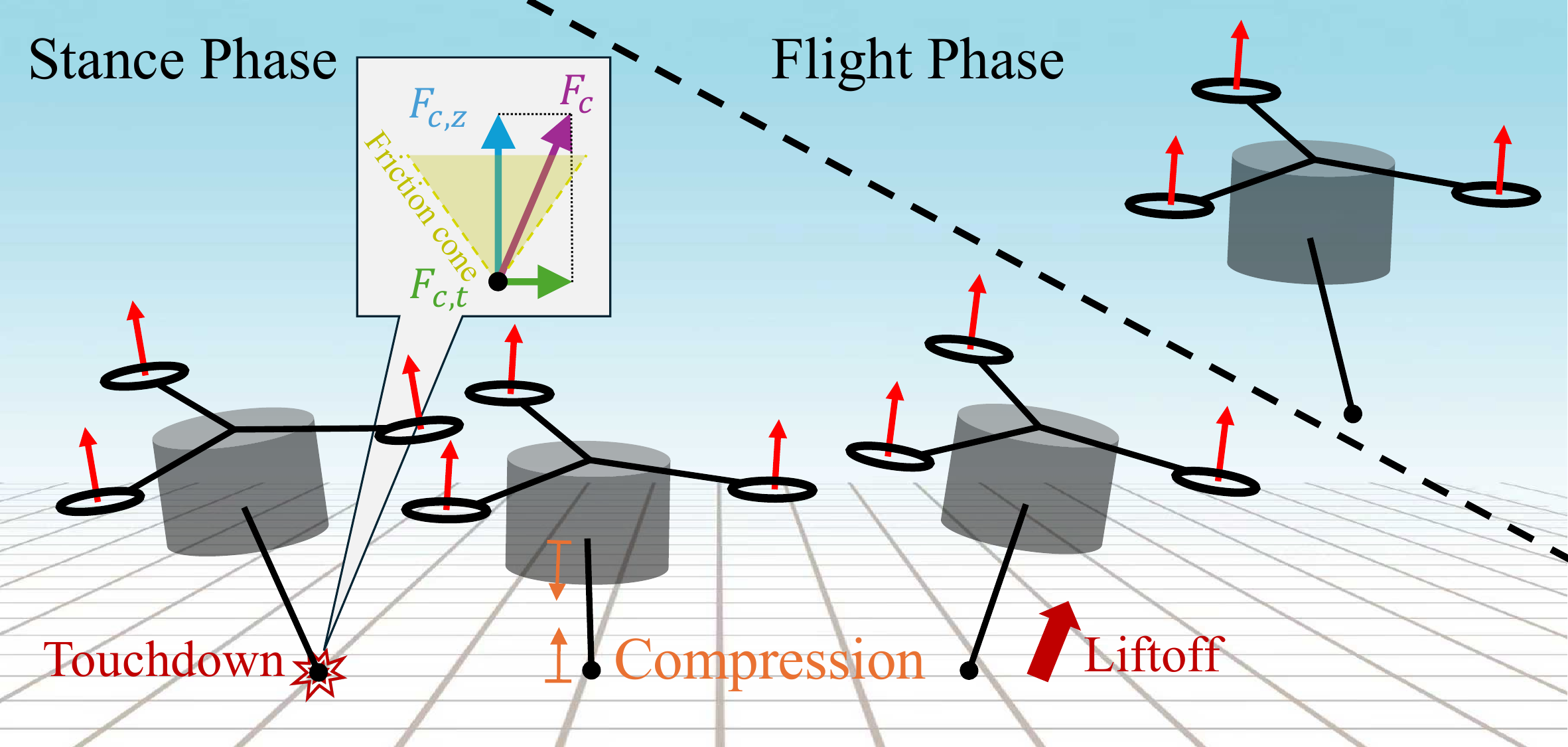}%
    }{%
        \fbox{\parbox{0.9\columnwidth}{\centering \small Phase Diagram Placeholder\\(missing: \texttt{figures/phase.pdf})}}%
    }
    \caption{Hopping cycle with touchdown and liftoff events. In stance, the leg contacts the ground; in flight, the tri-rotor stabilizes attitude.}
    \label{fig:phase}
\end{figure}

\subsection{Phase Detection and Vertical Height Control}
As illustrated in Fig.~\ref{fig:phase}, the controller detects the \emph{stance phase} and the \emph{flight phase} using a finite state machine driven by the leg length $l = \|\bm{p}_{foot}\|$: touchdown is detected when $l \le l_0 - \Delta_{td}$, and liftoff is detected when $l \ge l_0$, where $l_0$ is the nominal leg length and $\Delta_{td}$ is a hysteresis threshold for touchdown; its value is listed in Table~\ref{tab:parameters}.

During the stance phase, the vertical controller injects energy to reach a target apex CoM height $h_{des}$ while absorbing the landing impact. Here $h_{des}$ denotes the desired apex CoM height measured along the world vertical. We regulate the vertical motion using CoM height and vertical velocity feedback, and we add an energy compensation term based on the CoM height. The target energy corresponding to the desired apex height is defined as $E_{des} = mg h_{des}$.

The instantaneous vertical energy $E_{sys}$ is evaluated as
\begin{equation}
E_{sys} = \frac{1}{2} m \dot{p}_z^2 + mg\,p_z,
\label{eq:system_energy}
\end{equation}
where $p_z$ and $\dot{p}_z$ are the CoM vertical position and velocity estimated from encoder and IMU measurements during stance. At each touchdown, $p_z$ is re-anchored geometrically from the measured leg configuration and body orientation rather than integrated over time, so the estimate does not drift across hops. The commanded vertical contact force is then formulated as
\begin{equation}
\begin{aligned}
f_{c,z}^{des} &= \max\!\Big(0,\,k_z(h_{des} - p_z) - b_z \dot{p}_z + k_E(E_{des} - E_{sys})\Big) \\
&= \max\!\Big(0,\,(k_z + k_E mg)(h_{des} - p_z) \\
&\qquad\qquad - b_z \dot{p}_z - \tfrac{1}{2} k_E m\, \dot{p}_z^2\Big),
\end{aligned}
\label{eq:push_impedance_energy}
\end{equation}
where $k_z$ and $b_z$ are the vertical height and damping gains and $k_E$ is the energy feedback gain. The second line follows from substituting Eq.~\eqref{eq:system_energy} and clarifies the role of the energy term: rather than acting as an independent loop, it augments the effective height stiffness from $k_z$ to $k_z + k_E mg$ and adds a quadratic penalty $-\tfrac{1}{2} k_E m\, \dot{p}_z^2$ on the vertical kinetic energy. This kinetic-energy penalty is the genuinely new contribution of the energy term, modulating the vertical force during push-off and compensating for energy dissipated at impact.

\subsection{Centroidal Wrench and Flight Foot Placement}
To stabilize the spatial orientation against asymmetric disturbances, we first formulate a desired body-frame centroidal moment $\bm{\tau}_b^{des}$ using an SO(3) PD control law \cite{ref_lee2010}. Defining the orientation error as $\bm{e}_R = \tfrac{1}{2}(\bm{R}_{des}^\top\bm{R}-\bm{R}^\top\bm{R}_{des})^\vee$, where $(\cdot)^\vee$ denotes the vee map, the desired body-frame moment is:
\begin{equation}
\bm{\tau}_b^{des} = -\bm{K}_{R}\bm{e}_R + \bm{K}_{\omega}(\bm{\omega}^{des}-\bm{\omega}),
\label{eq:pd_attitude}
\end{equation}
where $\bm{R}_{des}$ is constructed from the desired roll and pitch together with the current yaw, so yaw remains uncontrolled, and $\bm{\omega}^{des}=\bm{0}$. The allocator uses the corresponding world-frame moment $\bm{\tau}^{des} = \bm{R}\bm{\tau}_b^{des}$.
Here $\bm{K}_{R}$ and $\bm{K}_{\omega}$ are diagonal with nonzero roll/pitch gains; the phase-dependent values are listed in Table~\ref{tab:parameters}. The commanded roll/pitch moment is saturated by $\tau_{\mathrm{rp,max}}^{\mathrm{stance}}$ in stance and by $\tau_{\mathrm{rp,max}}^{\mathrm{flight}}$ in flight, as listed in Table~\ref{tab:parameters}.

While the tri-rotor continuously regulates attitude, the underactuated horizontal motion is regulated at the stride level via a Raibert-style foot-placement policy. Let $\bm{v}_{xy}$ and $\bm{v}_{xy}^{des}$ denote the estimated and desired horizontal CoM velocities expressed in a body-aligned horizontal frame. Following a modified Raibert-style touchdown-placement rule, we compute the desired horizontal touchdown offset in the same frame as:
\begin{equation}
\bm{r}_{td,xy}^{des} = k_v \bm{v}_{xy} - k_r \bm{v}_{xy}^{des}.
\label{eq:s2s_raibert}
\end{equation}
Here, the term $k_v \bm{v}_{xy}$ provides velocity feedback for stabilizing the horizontal limit cycle, while the term $-k_r \bm{v}_{xy}^{des}$ biases the touchdown offset according to the commanded horizontal motion. Let $\bm{r}_{td,xy}^{des} = [x_{td}^{des}, y_{td}^{des}]^\top$. If $\|\bm{r}_{td,xy}^{des}\|_2 > l_0$, the horizontal target is clipped to the reachable disk before forming the full 3D foot-placement target in the body frame,
\[
\bm{p}_{foot}^{des} =
\begin{bmatrix}
x_{td}^{des} \\
y_{td}^{des} \\
-\sqrt{l_0^2 - (x_{td}^{des})^2 - (y_{td}^{des})^2}
\end{bmatrix},
\]
which is then tracked by the 3-RSR leg during the flight phase. Here the touchdown target is coordinated with an SRB-based centroidal controller and the aerial attitude action of the tri-rotor.
\begin{figure*}[t]
    \centering
    \includegraphics[width=0.95\linewidth]{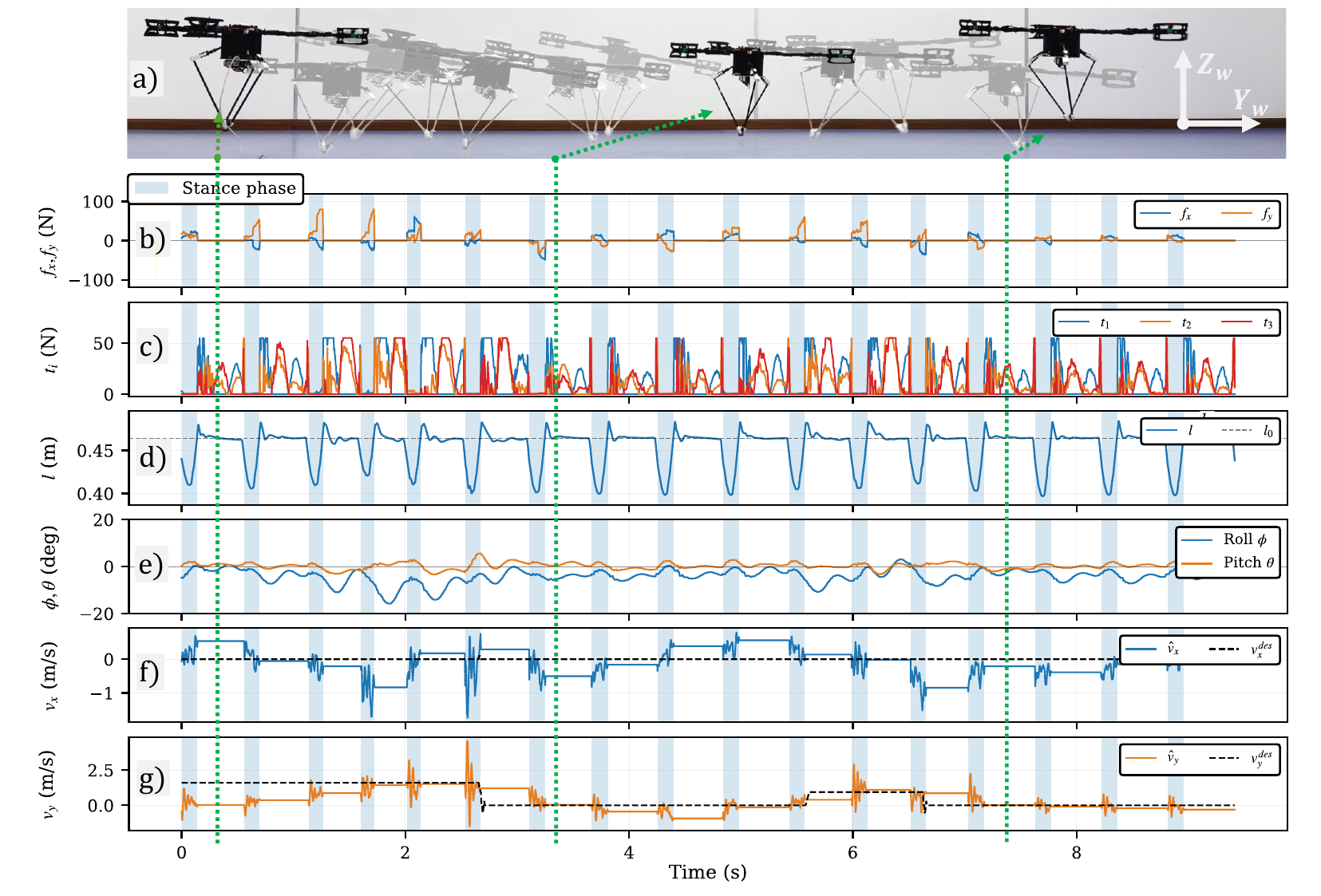}
    \caption{Indoor stable 3D hopping over a representative $9.4$\,s window with $17$ consecutive hops. The vertical green dotted lines mark the time instants corresponding to the key frames in (a). (a)~Photo sequence on flat indoor ground. (b)~Horizontal contact forces $f_x, f_y$. (c)~Propeller thrusts $t_{1,2,3}$. (d)~Leg length $l$; dashed line: $l_0$. (e)~Roll $\phi$ and pitch $\theta$. (f) and (g)~Horizontal velocity tracking, with measured velocity (solid) and commanded velocity $v_{x}^{des}, v_{y}^{des}$ (black dashed).}
    \label{fig:exp1}
\end{figure*}
\subsection{Hierarchical Force Allocation}
The desired centroidal wrench, parameterized by $f_{c,z}^{des}$ and $\bm{\tau}^{des}$, must be mapped to the physical actuators. Reflecting the leg-primary, tri-rotor-auxiliary role asymmetry, we resolve the leg wrench first and assign only the residual attitude moment to the propellers. This hierarchy enforces the leg-first priority by construction, so the propellers can only add to, never substitute for, the leg's contact action, avoiding the actuator trade-offs and contact-transition chattering of a tightly coupled joint optimization.

\subsubsection{Leg Force Mapping and Saturation in the Stance Phase}
Given the leg's role in generating stance contact forces, our allocation scheme prioritizes the 3-RSR mechanism. Importantly, the high-level moment $\bm{\tau}^{des}$ is not sent directly to the leg motors. Instead, we set $f_{c,z}^{cmd}=f_{c,z}^{des}$ and compute the horizontal contact force $\bm{f}_{c,xy}^{cmd}$ by solving a leg-priority constrained least-squares that tracks the desired roll/pitch moment while enforcing friction and motor-torque limits:
\begin{equation}
\begin{aligned}
\bm{f}_{c,xy}^{cmd} = \arg\min_{\bm{f}_{xy}} \ & \left\|(\bm{r}_c \times \bm{f}_c)_{rp} - (\bm{\tau}^{des})_{rp}\right\|_2^2 \\
\text{s.t. } \ & \bm{f}_c = [\bm{f}_{xy}^\top, f_{c,z}^{des}]^\top, \\
& \|\bm{f}_{xy}\|_2 \le \mu f_{c,z}^{des}, \\
 & \|\bm{J}^\top(\bm{q})(-\bm{R}^\top\bm{f}_c)\|_\infty \le \tau_{\mathrm{max}},
\end{aligned}
\label{eq:stance_force_qp}
\end{equation}
where $(\cdot)_{rp}$ extracts the roll and pitch components, and $\mu$ is the friction coefficient. This yields the feasible stance force $\bm{f}_c^{cmd}=[(\bm{f}_{c,xy}^{cmd})^\top, f_{c,z}^{des}]^\top$, which is then mapped to leg motor torques through Eq.~\eqref{eq:leg_mapping}.

\subsubsection{Propeller Residual Compensation in the Stance Phase}
However, under uneven terrain and external disturbances, the leg may not fully realize the restorative moment required within a single stance interval because the admissible stance force is limited by friction and motor-torque constraints. The residual attitude correction is therefore handled by the propellers.

The propellers act strictly as an on-demand residual compensator. After the leg-side allocation produces a feasible stance force $\bm{f}_c^{cmd}$, the corresponding stance moment allocated to the leg is
\begin{equation}
\bm{\tau}_{leg}^{alloc} = \bm{r}_c \times \bm{f}_c^{cmd},
\end{equation}
which is not an independent PD command, but the moment actually assigned to the leg by the feasible contact-force allocation. The propeller command then compensates only the residual part left after this feasible leg allocation:
\begin{equation}
\bm{\tau}_{res}^{cmd} = \bm{\tau}^{des} - \bm{\tau}_{leg}^{alloc}.
\label{eq:tau_residual}
\end{equation}
Thus, the first layer defines the desired net body moment $\bm{\tau}^{des}$, while the second layer lets the leg realize as much of that demand as possible within its feasible region and assigns only the unallocated residual part to the tri-rotor. A nonzero baseline collective thrust $T_{base}$ is maintained so that the propellers remain responsive while generating the residual roll and pitch torque. Given $\bm{\tau}_{res}^{cmd}$, the thrust deviations are obtained from
\begin{equation}
\begin{aligned}
\delta\bm{t}^{cmd} = \arg\min_{\delta\bm{t}} \ & \left\|(\bm{M}_t\delta\bm{t})_{rp} - (\bm{\tau}_{res}^{cmd})_{rp}\right\|_2^2 \\
\text{s.t. } \ & -\tfrac{T_{base}}{3} \le \delta t_i \le t_{\mathrm{max}} - \tfrac{T_{base}}{3}, \\
& \mathbf{1}_3^\top\delta\bm{t} \le T_{\mathrm{max}} - T_{base},
\end{aligned}
\label{eq:prop_residual_lsq}
\end{equation}
where $\bm{M}_t = [\bm{r}_1 \times \bm{d}_t \ \ \bm{r}_2 \times \bm{d}_t \ \ \bm{r}_3 \times \bm{d}_t] \in \mathbb{R}^{3 \times 3}$ is the world-frame tri-rotor moment map, $T_{\mathrm{max}}$ is the total available thrust limit of the tri-rotor, and the final thrust command is $\bm{t}^{cmd} = \tfrac{T_{base}}{3}\mathbf{1}_3 + \delta\bm{t}^{cmd}$.

\subsubsection{Aerial Attitude Regulation in the Flight Phase}
Upon liftoff, the contact wrench vanishes ($\bm{f}_c = \bm{0}$). The tri-rotor then becomes the primary attitude actuator and tracks the roll and pitch components of $\bm{\tau}^{des}$ while maintaining the baseline collective thrust $T_{base}$. Meanwhile, the leg tracks the foot-placement target for the next touchdown. This provides continuous attitude authority in the air and helps ensure that the robot returns to stance with a favorable body orientation.

\section{EXPERIMENTAL VALIDATION}

The experiments evaluate the proposed HFA framework in three scenarios: indoor velocity tracking, outdoor hopping across an uneven grass-to-wood transition, and push recovery under impulsive disturbances. These tests probe phase-dependent coordination of the leg and propellers, robustness to unmodeled contact variation, and recovery from severe attitude and horizontal-velocity perturbations. In particular, Fig.~\ref{fig:exp1} provides the clearest quantitative view of the phase-dependent coordination between leg contact action and propeller assistance within repeated hopping cycles. All experiments use identical controller parameters listed in Table~\ref{tab:parameters} and the same hardware/software stack described in Section~II, with velocity commands published via a handheld gamepad.
\subsection{Indoor Stable 3D Hopping}
Figure~\ref{fig:exp1} shows a representative $9.4$\,s window of indoor hopping under piecewise forward and lateral velocity commands. In particular, the lateral command contains two acceleration-and-stop segments, with peak commanded speeds of approximately $1.6$\,m/s and $0.94$\,m/s. The leg reaches a peak stance compression of approximately $67.3$\,mm relative to the nominal length $l_0$, indicating stable vertical energy regulation, while the body attitude remains bounded throughout the sequence (roll approximately $[-15.8^{\circ},\,3.1^{\circ}]$, pitch $[-3.3^{\circ},\,5.6^{\circ}]$) despite repeated impacts and command reversals.

\begin{figure*}[t]
    \centering
    \includegraphics[width=0.95\linewidth]{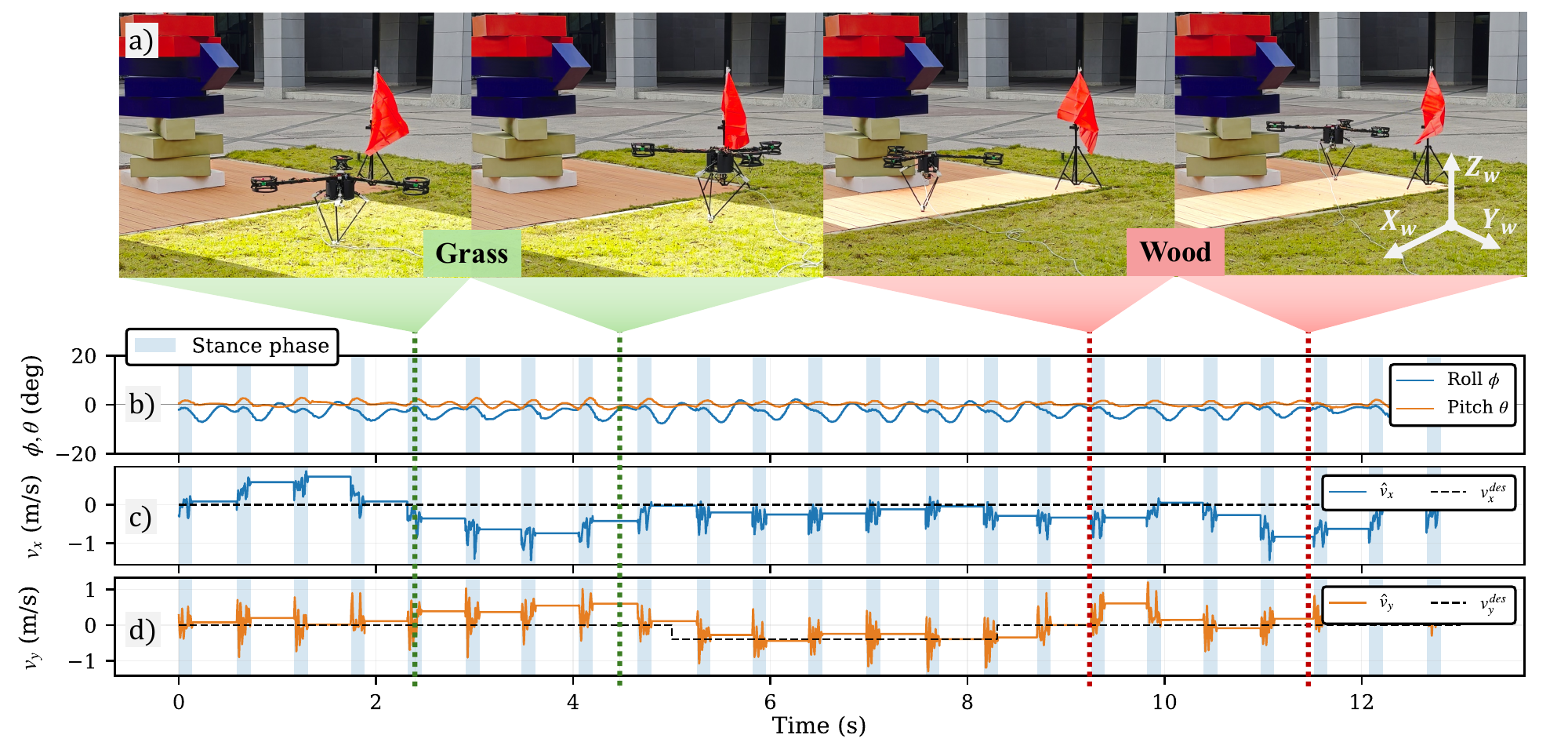}
    \caption{Stable 3D hopping on uneven outdoor terrain over a representative $13$\,s window with $23$ consecutive hops. (a)~Photo sequence across the grass-to-wood transition. (b)~Attitude $\phi, \theta$; blue shading indicates stance. (c) and (d)~Horizontal velocity tracking.}
    \label{fig:exp2}
\end{figure*}
\begin{figure*}[t]
    \centering
    \includegraphics[width=0.95\linewidth]{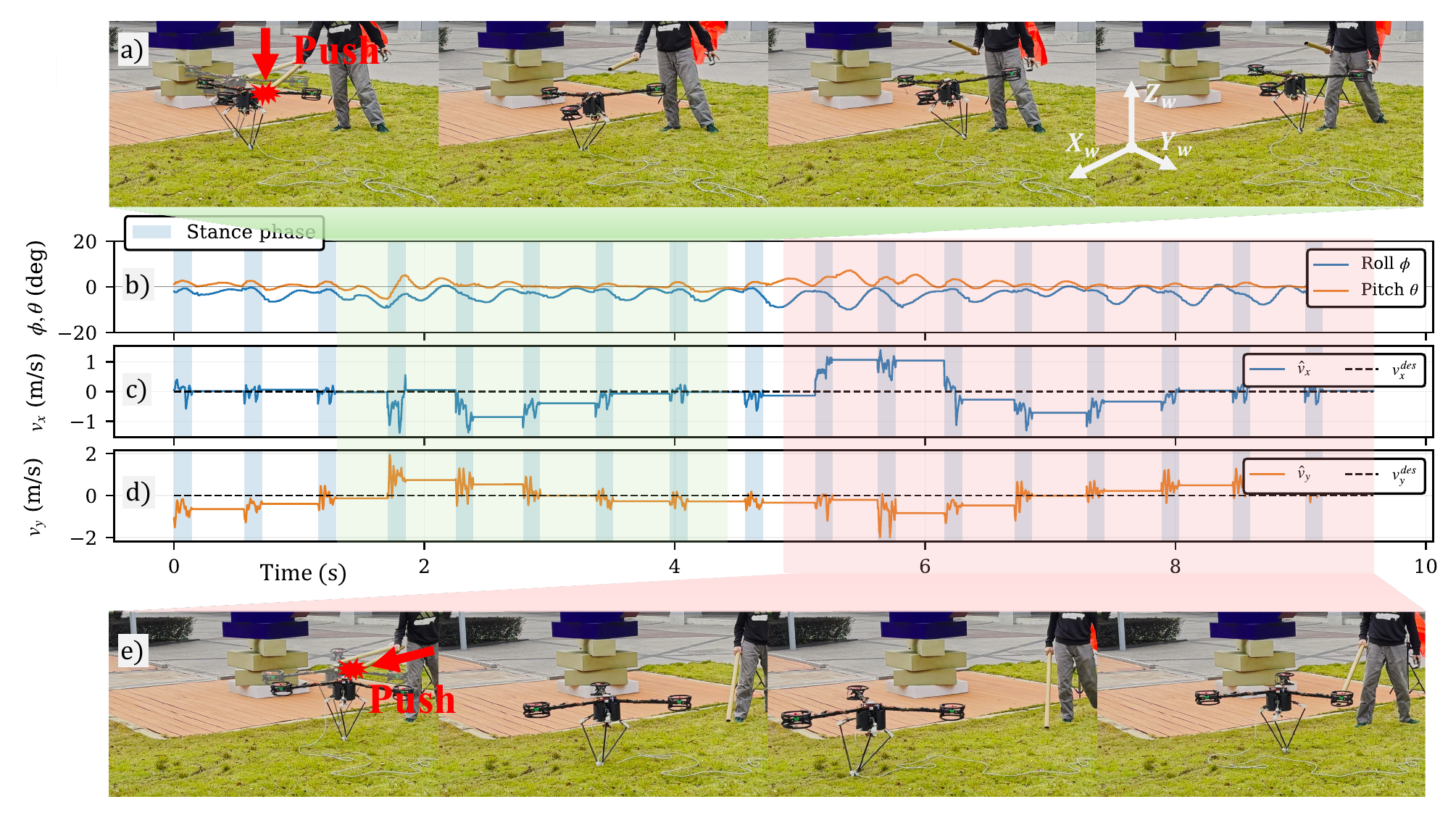}
    \caption{Push recovery during outdoor hopping in a representative $9.6$\,s trial with $17$ hops and two pushes. (a)~First push at approximately $t{=}1.5$\,s. (b)~Attitude $\phi, \theta$; blue shading indicates stance. (c) and (d)~Horizontal velocity tracking. (e)~Second push at approximately $t{=}4.9$\,s, approximately along $+x$.}
    \label{fig:exp3}
\end{figure*}
Figure~\ref{fig:exp1}(b) and (c) show the phase-dependent action of HFA: commanded horizontal contact forces occur primarily during stance, whereas propeller thrust modulation increases after liftoff. The horizontal velocity traces in Fig.~\ref{fig:exp1}(f) and (g) are consistent with the Raibert-style foot-placement policy in Eq.~\eqref{eq:s2s_raibert}, showing stride-to-stride recovery of horizontal motion without destabilizing the vertical limit cycle.

\subsection{Outdoor Hopping}
As shown in Fig.~\ref{fig:exp2}, the robot hops outdoors across a grass-to-wood transition under light wind. The plotted interval is a representative $13$\,s window containing $23$ consecutive hops, while the robot continues hopping beyond the displayed segment. Without terrain perception, contact-timing adaptation, or gain retuning, the robot maintains bounded attitude motion throughout the transition. In particular, the roll and pitch traces remain within approximately $[-7.7^{\circ},\,2.2^{\circ}]$ and $[-2.2^{\circ},\,2.8^{\circ}]$, with standard deviations of $\sigma_\phi{=}2.1^{\circ}$ and $\sigma_\theta{=}1.0^{\circ}$, respectively. The attitude deviations increase only moderately after uneven impacts and decay again within 1 to 2 hops, while the horizontal velocity remains close to the commanded references. These observations indicate that the phase-dependent HFA structure tolerates substantial unmodeled variations in contact compliance and terrain transition without requiring controller retuning. This outdoor behavior is also consistent with Fig.~\ref{fig:teaser}, which illustrates the benefit of propeller assistance on grassy terrain.

\subsection{Push Recovery}
To evaluate disturbance recovery, we applied two impulsive pushes during outdoor hopping (Fig.~\ref{fig:exp3}). The first push (key frame at approximately $t{=}1.5$\,s in Fig.~\ref{fig:exp3}(a)), applied downward on a rotor arm, induces a pronounced attitude disturbance; the propellers recover roll/pitch during the subsequent flight phase and return the robot to a stable hopping cycle within 1 to 2 hops (Fig.~\ref{fig:exp3}(b)), consistent with the residual compensation mechanism in Eq.~\eqref{eq:tau_residual}.

The second push (key frame at approximately $t{=}4.9$\,s in Fig.~\ref{fig:exp3}(e)), applied laterally to the main body approximately along the $+x$ direction, produces a horizontal velocity spike of approximately $1.4$\,m/s in $v_x$ (Fig.~\ref{fig:exp3}(c) and (d)); the Raibert-style foot-placement policy dissipates the injected momentum and restores velocity tracking within three hops without destabilizing the vertical motion. Together, these tests demonstrate that the controller distributes recovery across components with distinct roles: residual attitude compensation by the propellers and stride-level horizontal recovery by foot placement.

\section{CONCLUSION AND FUTURE WORK}
This paper presented Pro-OMEGA2, a propeller-assisted 3D hopping robot that integrates an active 3-RSR parallel leg and a trunk-mounted tri-rotor unit.

We proposed an SRB-based HFA framework that preserves leg-generated stance contact forces as the primary locomotion action while using the tri-rotor for auxiliary attitude regulation, including residual attitude compensation in stance and continuous roll/pitch control during flight. Real-robot experiments demonstrate sustained indoor and outdoor 3D hopping, including abrupt start and stop velocity commands, grass-to-wood terrain transitions, and impulsive push recovery. Future work will move beyond phase-decoupled allocation toward predictive coupled control by integrating SRB-based MPC with QP-based wrench allocation. We also plan to incorporate full-order dynamics and learning-based models to better capture contact uncertainty and the coupling between the leg and propellers.

\bibliographystyle{IEEEtran}
\bibliography{references}

\end{document}